*A.B. Kutuzov*
*akutuzov72@gmail.com*
*Tyumen State University*

# Using descriptive mark-up to formalize translation quality assessment

## Contents

## *Abstract*

The paper deals with using descriptive mark-up to emphasize translation mistakes. The author postulates the necessity to develop a standard and formal XML-based way of describing translation mistakes. It is considered to be important for achieving impersonal translation quality assessment. Marked-up translations can be used in corpus translation studies; moreover, automatic translation assessment based on marked-up mistakes is possible. The paper concludes with setting up guidelines for further activity within the described field.

## *1. Introduction*

The problem of translation quality assessment is a rather old one. It seems that each and every translation school has developed or is developing its own criteria for impersonal assessment of translation. These criteria are employed both in real translation jobs to define quality of a given translation (and possibly to establish proper payment) and in translation didactics when we asses training translations done by students.

It is not a secret that criteria of translation quality are often subjective and depend on the assessor - whether translation project manager or teacher. This state of things brought about a lot of efforts to develop an impersonal system of translation quality assessment which would include the definition of adequate (or equivalent) translation, classification of translation mistakes by their types and value. For example, cf. [Паршин, 2002, Богатырёв, 2004, Сдобников, 2007, House, 1997, Hönig, 1997, Munday, 2001]

It is a disputable issue, which of these systems is more impersonal and fits the needs of translation industry or translation didactics better. We would like to pay attention to the 'dark side' of this problem.

## 2. Making translation quality description formal

It seems that discussions on objectiveness of translation quality assessment systems constantly evade the problem of formal representation of such systems. Even when a teacher or a manager (we will further refer to this person as 'critic') has developed a system of assessment good enough for him/her and for his/her clients, there is still the need to apply this theoretical system to real translations. Moreover, in modern computerized world you have to store data on translation mistakes and to make this data easily accessible.

It is not very difficult to store source and target texts together. Bilingual and parallel corpora received much attention since at least 1970 (for this see [Kennedy, 1998]).) Further development of this idea was the standard of translation memory files exchange - TMX (Translation Memory eXchange). It appeared in 1998[1].

But TMX is definitely not sufficient for **assessment** of translation and for storing information on translation mistakes. The very concept of translation memory implicitly presupposes that the text is translated correctly. As a result it lacks means to deal with mistakes. But for critical translation studies we need to fix the fact of **incorrect**, inadequate translation and to point at the nature of mistake. Moreover, it should be done in a formal way, subject to automated processing. This will allow to store and transmit mistake information easily.

Thus, we need to develop a formal method of marking up mistakes in the text of translation. This method must be flexible in order to adapt to translation assessment criteria adopted in different bodies and organizations. Naturally, translation texts must be in digital form, else computers won't be able to process them.

## 3. Employing linguistic mark-up to emphasize translation mistakes

The key word of the previous paragraph is **mark-up**. It is commonly known that mark-up is the use of embedded codes known as tags to describe a document structure or to embed instructions that can be used by document management tools. [Raymond et al, 1992] Thus, mark-up gives information about the text by using some elements placed within the text, but not being its part. A well-known example of mark-up is punctuation. When we read a text aloud, the dot sign tells us that we should break an intonation segment and perhaps to make a pause.

The authors of classical «Markup Systems and the Future of Scholarly Text Processing» enumerate five types of mark-up:
1. Punctuational

---

1 Current TMX specifications are available on-line at
http://www.lisa.org/fileadmin/standards/tmx1.4/tmx.htm.

2. Presentational
3. Procedural
4. Descriptive
5. Referential
6. Meta mark-up
   [Coombs et al, 1987]

For our needs it is crucial to separate procedural, presentational and descriptive mark-up. To cut it short, procedural mark-up describes what you have to **do** with elements of text, presentational mark-up describes how they should **look** and descriptive mark-up describes what these elements **are**.

Further goes an example of one and the same piece of text represented in different mark-ups. Elements of mark-up (tags) are given in italics, coloured red and enclosed within angle brackets. Tags beginning with a slash (*</>)* are end tags, that is, they signal the end of what was declared by the start tag.

| Type of mark-up | Text |
|---|---|
| **Without mark-up** | **Translation mistakes**<br>Translation mistakes can be form mistakes or content mistakes. |
| **Procedural mark-up** | *<Step 20 points from the left margin><choose bold letter type>*Translation mistakes*<go to the next line><choose regular letter type>* Translation mistakes can be form mistakes or content mistakes. |
| **Presentational mark-up** | *<centre alignment><bold font face>*Translation mistakes*</bold font face></centre alignment> <left alignment><regular font face>*Translation mistakes can be form mistakes or content mistakes*</regular font face></left alignment>* |
| **Descriptive mark-up** | *<header>*Translation mistakes*</header> <main text>*Translation mistakes can be form mistakes or content mistakes*</main text>* |

Computer will display these four texts (almost) similarly, but hey are marked up in accordance with different principles. Procedural mark-up gives rendering software firm commands and tells nothing about the nature of text elements or even how exactly they will look like. Presentational mark-up belongs to higher abstraction level; it tells what kind of design must be applied to this or that element of text. This type of mark-up leaves aside any commands to apply this design (we presuppose that our rendering software is responsible for that). But, similarly to procedural mark-up, it does not tell anything about the nature of text elements[2]. Finally, descriptive mark-up describes the **structural** role of an element (for example, "header", "list" or "'element of bibliography") and presupposes that our rendering software will choose the right correspondent variant of design for this element.

---

2  An example of presentational mark-up is HTML (Hyper-Text Markup Language).

I seems that formal description of translation mistakes should be done with descriptive mark-up. Translation critic must concentrate on the **content** of his/hers corrections (for example, what is the type of the mistake), not on their **look** (for example, font face and size). Exact look of marked up translation should be constructed only when it is being rendered and it should depend on the settings of rendering software.

Another feature of descriptive mark-up is the possibility of **multiple representations** of one and the same text. It means that depending on our tasks we can interpret and render tags in different ways (or do not render them at all). For example, a critic might want to look at the translation text with emphasized content mistakes or form mistakes or only critical mistakes. As long as descriptive mark-up does not deal with precise layout and look, it allows a critic to simply point at the types of mistakes in the text. Thus, it builds the basis for any possible representation.

Similar system (though not for translation but for essay teaching purposes) is described in «Computer Supported Proofreading Exercise in a Networked Writing Classroom» [Hiroaki et al, 1999]. Interestingly, we did not manage to find researches devoted to translation mark-up neither in Russian nor in foreign translation studies; this fact makes the issue even more topical.

## 3.1 Why XML?

Today's standard of descriptive mark-up is **XML** (eXtensible Markup Language) — platform, software and hardware independent means of transmitting data. XML is a heir of another mark-up language - SGML. XML is used to store any structured data, including texts. In fact, it is a set of syntactic rules to describe data structure. XML itself does not give any list of functional elements (tags). Instead, it allows you to define your own sets of tags, using so called XML Schemata to describe them. It is like creating dialects. Schemata describe tag vocabularies for these dialects and correct way of their usage. With the help of these tags you can mark up structure, content and semantics of an XML document.

In order to mark up textual data (corpora) several western universities[3] developed a scheme describing which text parameters are to be marked up. It employs XML and is called Text Encoding Initiative Guidelines (TEI Guidelines[4]). It is a list of various features of text which you can code, mark-up and index. For example, the system enumerates various types of corrections, misprints, quotations, foreign words, etc. As of now, most text corpora projects (including widely known British National Corpus, BNC) try to more or less follow the guidelines of TEI.

Thus, in order to mark up a translation text we have to develop an XML schema (dialect) which will include a set of tags classifying translation mistakes and a description of their correct usage. This schema must keep in mind TEI guidelines to meet international requirements for linguistic mark-up. Besides, the schema must be flexible and allow to change itself in accordance with classification of translation mistakes adopted within this or that organization.

---

3  Oxford, Brown, Virginia and some others
4  http://www.tei-c.org/Guidelines/index.xml

XML allows to employ a host of already developed software for dealing with marked-up documents and gives possibility to re-use existing textual corpora.

## 4. Modelling use of translation mark-up

We see XML within translation quality assessment in the following way (let's take translation didactics as an example).

A critic receives a translation done by a student and checks it. While checking, he or she marks mistakes with corresponding tags (defined within the current schema). Tag placement can be done by hand (rather labour-intensive) or with the help of a computer - as simple as selecting a mistake and pressing a button in graphic user interface (of course it will demand developing such an interface first). The result is a marked-up text containing information  about translation mistakes. Corpus of such texts can be used for virtually any research related to translation process - you can easily find, say, "all cases of content mistakes with the word *magazine*".  In the future one can create huge marked-up corpora of students' translations.

As we noted earlier, a text with descriptive mark-up can be rendered in different representations, allowing to look at translation from different "points of view" without changing anything in the text itself and in mark-up. Naturally, we'll have to develop tunable rendering software or (preferably) to use the existing browsers like Internet Explorer or Mozilla Firefox. Modern browsers are already able to deal with XML, we just need to give them a style sheet (CSS) defining how this of that type of translation mistakes should be formatted.

Thus, we get a method of formal description and further display of translation mistakes. But this is not the only application of XML mark-up in translation.

## 4.1 Automated translation assessment

Riccardo Schiaffino points that «A correct translation is a translation with no errors or where total error points result in a Translation Quality Index above the desired threshold» [Schiaffino, Zearo, 2003]. One can see the introduction of a notion of Translation Quality Index, TQI, which can be higher or lower than some threshold stated by a client or a teacher. The notion of TQI was in fact introduced in 1995 by LISA association and defines translation quality in percent based on the quantity of "error points". The quantity of error points assigned to each mistake depends on the weights of mistake types. These weights are also stated by the client. For example, in professional or scientific texts we value standardization and term consistency, that's why we can assign more error points to term mistakes in texts like this. Once we calculated the TQI, we can give a mark to translation - of course in accordance with the standards adopted in the current organization or effective in the current situation.

Usage of XML mark-up gives possibility to automatize calculation of TQI and even giving the final mark. The task of teacher (checker) is limited to marking up mistakes; the assessment is done automatically basing on given weights of mistakes and TQI thresholds. Thus, this method allows to divide detection and classification of mistakes (partially subjective process) and translation quality assessment itself (purely formal and impersonal process). In principle, these activities can be done by different persons or even departments. Talking about translation didactics, it would be useful to define different TQI thresholds for different years of studying and different control points (common test, mid-term exam, final exam, etc). For example, we can give a "good" mark for translation with TQI=90% if it is done by a freshman, but only "satisfactory" in case of a senior student.

## 5. Example of marked-up translation.

Further we give an example of applying XML mark-up to translation mistakes. Before marking up we have to (at least briefly) define the schema of mark-up; in order to do this we have to choose one of the existing classification of mistakes. Let's take classification of Nizhniy Novgorod tradition (of course, simplified). According to this conception, translation mistakes can be of content (distortion of factual or communicative information), of form (violation of target language standards) or of style (incorrect transmission of stylistic devices) [Сдобников, 2007]. Thus, in our example we will put mistakes into one of these types; besides we will give 'mistake value' - it can be minor, major or critical.

We use tags as indicators of mistakes. Each XML tag has its name (or type) and a set of attributes (it can be empty). Tag describes some part of the text (it can be empty as well)[Yeates, 2006].

Let's assign the following names for mistake tags: *<content_mistake>* for content mistakes, *<form_mistake>* for form mistakes and *<style_mistake>* for style mistakes. Mistake value will be indicated by attribute weight of the corresponding tag — minor, major or critical. For example, a critical content mistake can be indicated like this: *<content_mistake weight='critical'>* here we have incorrect translation*</content_mistake>*.

Let's apply this XML dialect to a real student's translation from English to Russian.

Source:

*There are two possible approaches to automating the translation process. Machine translation has been a "holy grail" of the IT industry for more than 40 years. There have been significant advances in language technology over this period and we all benefit from these on a day to day basis when we use spelling and grammar checkers and ever more sophisticated search engines.*

Target:

*Два способа переводить «ничего» не делая. Уже на протяжении сорока лет машинный или автоматический перевод остаётся «голубой мечтой» сферы информационных технологий. И хотя результаты машинного перевода оставляют желать лучшего, данная технология анализа языка*

*претерпела значительные улучшения, позволяющие нам уже сегодня пользоваться проверкой правописания и грамматики, и даже такими сложными инструментами как поисковые машины.*

Having in mind the tags we defined earlier, this text can be marked up like this:

*Два способа переводить <style_mistake weight='critical'>«ничего» не делая</style_mistake>. Уже на протяжении сорока лет машинный или автоматический перевод остаётся «голубой мечтой» сферы информационных технологий. <content_mistake weight='minor'>И хотя результаты машинного перевода оставляют желать лучшего</content_mistake>, данная технология анализа языка претерпела значительные улучшения, позволяющие нам уже сегодня пользоваться проверкой правописания и грамматики, и даже такими сложными инструментами как поисковые машины.*

What is the sense of our mark-up? We pointed that in the first sentence the student translated the stylistic colouring of the phrase incorrectly, because he replaced scientific term *automating* with 'layman' evaluative expression "ничего не делая". We considered this to be a serious distortion of the communicative intention of the text, thus we marked this style mistake as critical (*weight='critical'*). The second mistake is the introduction of phrase *«И хотя результаты машинного перевода оставляют желать лучшего»*, which we do not find in the source sentence. This is a content mistake called "addition of information". However, we know that further in the text we can actually find this utterance about automatic translation (so it does not contradict the general sense of the text). That's why we assigned this mistake minor value (*weight='minor'*).

Now we can render the marked up text in various representations. Say, if a critic wants to visually show the value of translation mistakes, he or she can tune rendering software so that it emphasized minor mistakes with green background and critical ones with red background:

*Два способа переводить «ничего» не делая. Уже на протяжении сорока лет машинный или автоматический перевод остаётся «голубой мечтой» сферы информационных технологий. И хотя результаты машинного перевода оставляют желать лучшего, данная технология анализа языка претерпела значительные улучшения, позволяющие нам уже сегодня пользоваться проверкой правописания и грамматики, и даже такими сложными инструментами как поисковые машины.*

Additionally, all mistakes are given in bold. As one can see type of mistake is ignored in this representation. This is just one of many possible representations of our marked-up translation. We should keep in mind that the text itself remains unchanged, because descriptive mark-up describes only the meanings of text elements, not the way of their rendering.

Moreover, basing on this text we can calculate TQI and give a mark. For example, if our assessment standards assign 3 error points for each critical mistake and 1 error point for each minor mistake (assuming that the type of mistake does not matter), then the total quantity of error points in this translation equals to 4 points in 53 word, which gives us TQI = $100-(\frac{4}{53}\times100)=93$ percent. Depending of the adopted TQI thresholds this number is transformed into final mark, say, "good" if we stated that this mark is given if TQI is between 85% and 95%.

## 6. Conclusion

We pointed at the importance and prospectiveness of applying XML mark-up to incorrect translations. Translation science should develop a formal method of describing translation quality; descriptive mark-up fits good for this task. In accordance with local criteria and standards of translation assessment one can define different sets of XML tags describing translation mistakes. This sets must respect general syntactic rules defined in the schema of translation mark-up. We are in the beginning of the development of such a schema. However, it is already clear that this schema must keep in mind TEI guidelines to be portable and flexible. Then we have to develop software (or plug-ins for existing software) which will facilitate marking-up and rendering of marked-up texts. This task is eased by a large set of existing software able to deal with XML.

Marked-up translation can be assessed in an automated way, based on the marked up mistakes, their types and value. For this we use the notion of Translation Quality Index. It allows to separate detection of mistakes from assessment itself. It will lead to more impersonal marks.

We visualize the following main tasks in this field for the nearest future:

- composing detailed and TEI-compatible XML Schema for marking up translation mistakes;
- formal description of the most popular classifications of translation mistakes and defining dialects of root Schema for each of these classifications;
- development of software for marking up translations;
- development of tunable software for rendering of marked-up translations;
- formal description of criteria of translation assessment depending on the situation and demands of the client (teacher);
- development of software for automated assessment of marked-up translations.